\documentclass{article}
\usepackage{spconf,amsmath,graphicx,epstopdf,diagbox,amsfonts}
\usepackage{tabu}
\usepackage{subfigure}
\begin{document}\sloppy

% Example definitions.
% --------------------
\def\x{{\mathbf x}}
\def\L{{\cal L}}

% Title.
% ------
\title{Improving object detection with region similarity learning}

\name{FENG GAO$^{*,1}$, YIHANG LOU$^{*,1,2}$, YAN BAI$^{1,2}$, SHIQI WANG$^{3}$, TIEJUN HUANG$^{1}$, LING-YU DUAN$^{1}$ \thanks{$^{*}$FENG GAO and YIHANG LOU are joint first authors. }\thanks{LING-YU DUAN is the corresponding author.}}

\address{
$^{1}$National Engineering Lab for Video Technology, Peking University, Beijing, China \\
$^{2}$SECE of Shenzhen Graduate School, Peking University, Shenzhen, China \\
$^{3}$Rapid-Rich Object Search Laboratory, Nanyang Technological University, Singapore\\}
\maketitle
%Extensive experiments have shown the superiority of the proposed object detector,
\begin{abstract}
Object detection aims to identify instances of semantic objects of a certain class in images or videos. The success of state-of-the-art approaches is attributed to the significant progress of object proposal and convolutional neural networks (CNNs). Most promising detectors involve multi-task learning with an optimization objective of softmax loss and regression loss. The first is for multi-class categorization, while the latter is for improving localization accuracy. However, few of them attempt to further investigate the hardness of distinguishing different sorts of distracting background regions (\textit{i.e.}, negatives) from true object regions (\textit{i.e.}, positives). To improve the performance of classifying positive object regions vs. a variety of negative background regions, we propose to incorporate triplet embedding into learning objective. The triplet units are formed by assigning each negative region to a meaningful object class and establishing class-specific negatives, followed by triplets construction. Over the benchmark PASCAL VOC 2007, the proposed triplet embedding has improved the performance of well-known FastRCNN model with a mAP gain of 2.1\%. In particular, the state-of-the-art approach OHEM can benefit from the triplet embedding and has achieved a mAP improvement of 1.2\%.
\end{abstract}
\vspace{-10pt}
\section{Introduction}
\label{sec:intro}

%Recently, there has been an exponential increase in the demand for object detection, which aims at localizing target objects within images.
Object detectors {\cite{multiregion,girshick2014rich,girshick2015fast,ren2015faster,he2014spatial,song2011contextualizing,insideoutside} are normally trained as classifiers to recognize the objects contained in candidate boxes. For example, most traditional detectors like deformable parts models (DPM) \cite{DPM} work on sliding window techniques where the classifiers are run over the entire image at specified interval locations. More recent detectors like R-CNN \cite{girshick2014rich} employ a region proposal method to produce potential bounding boxes in an image and perform classification over these boxes. Object proposals can significantly reduce unnecessary predictions, and alleviate the negative effects of distracting background on region classification.
As there exist hard negative regions (their intersection over union (IoU) with any groundtruth object region $< 0.5$) involving part of background as well as parts of objects, as illustrated in Fig.~\ref{fig:metric}. How to accurately discriminate between such hard negative background regions and positive object regions ($IoU \ge 0.5$) is expected to improve the performance of object detectors.

\begin{figure}[]
\centering
\vspace*{-5pt}
\includegraphics[width=1\linewidth]{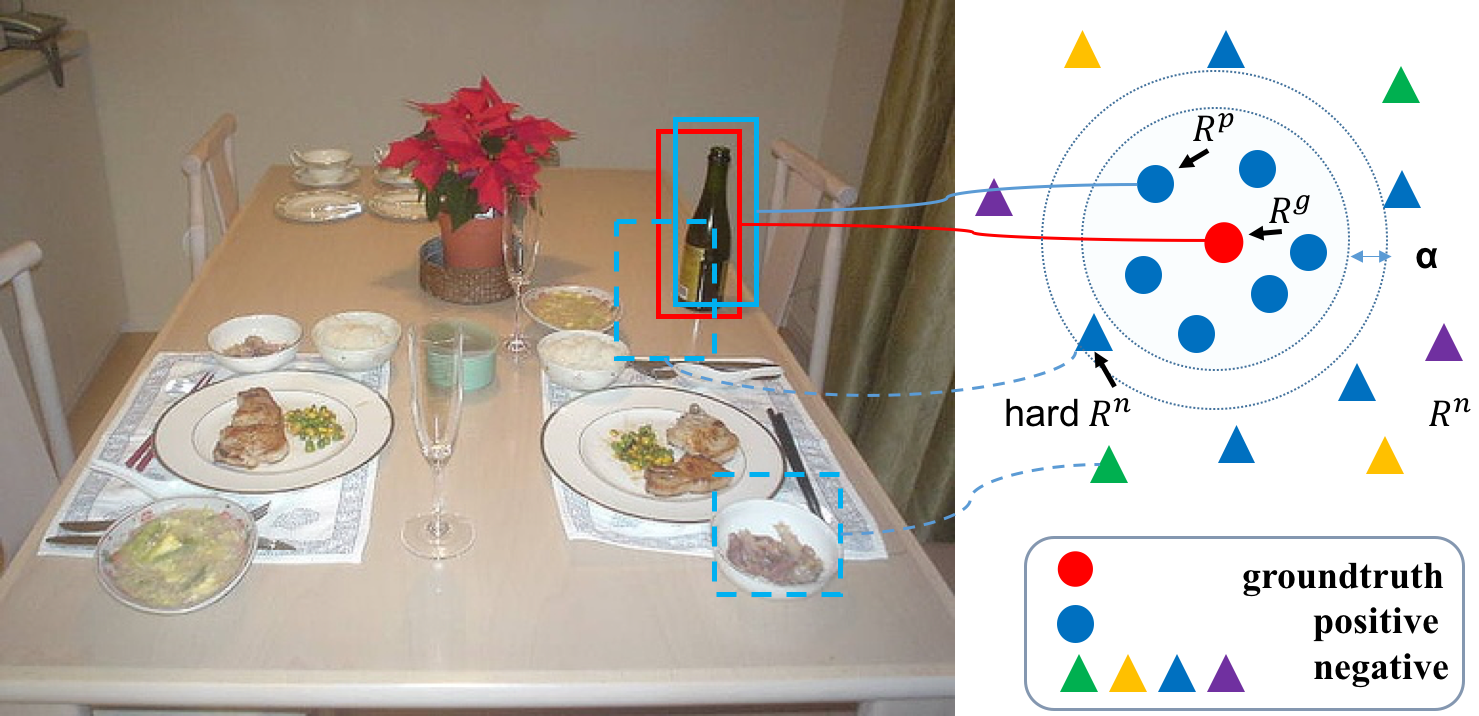}
\caption{Illustration of RoI feature distribution of the groundtruth, positives and negatives. The proposed approach aims to improve the RoI classification performance by dealing with hard negatives via similarity distance learning.}
\label{fig:metric}
\vspace*{-15pt}
\end{figure}

To improve the performance of detectors, many research efforts have been made to strengthen the capability of distinguishing positive regions and negative regions. In FastRCNN \cite{girshick2015fast}, many hyperparameters are introduced for efficient learning, \textit{e.g.}, the thresholds to define foreground RoIs (regions of interest) and background RoIs, the sampling ratio of positive (foreground RoIs) and negative samples (background RoIs) in mini-batch stochastic gradient descent (SGD) optimization, etc. Li \textit{et al}. \cite{attentivecontext} proposed to use LSTM cells \cite{hochreiter1997long} to capture local context information of proposal boxes and global context information of entire images to strengthen the discrimination ability of RoI's feature. In \cite{li2015scale}, Li \textit{et al.} took advantages of multiple subnetworks' output to deal with large scale changes. In \cite{insideoutside}, Bell \textit{et al.} introduced the Inside-Outside Net to capture multi-scale representation and incorporated context via spatial recurrent units.

\begin{figure*}
\vspace{-5pt}
\begin{minipage}[t]{0.24\linewidth}
  \centering
\centerline{\includegraphics[width=0.97\linewidth]{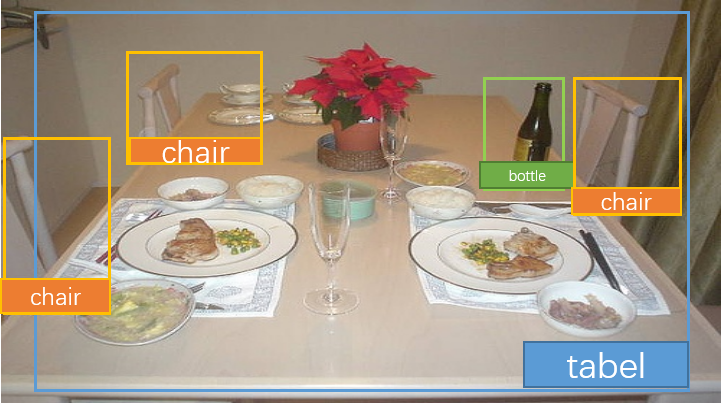}}
  \centerline{(a)}
\end{minipage}
\begin{minipage}[t]{0.26\linewidth}
\centering
\centerline{\includegraphics[width=1\linewidth]{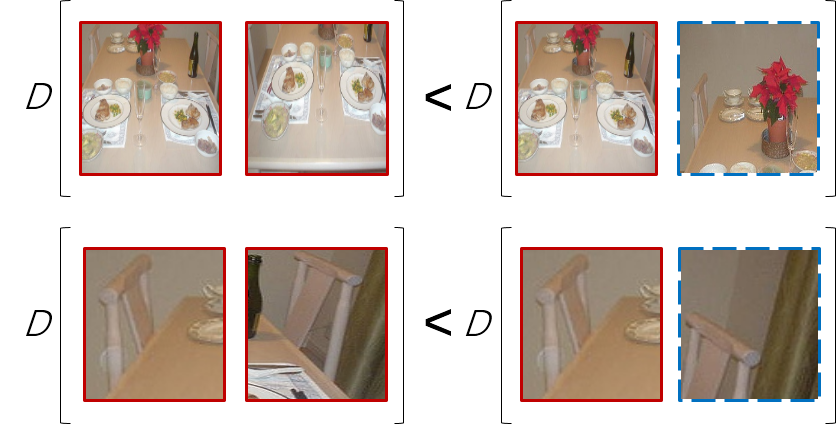}}
  \centerline{(b)}
\end{minipage}
\begin{minipage}[t]{0.49\linewidth}
\centering
\centerline{\includegraphics[width=0.97\linewidth]{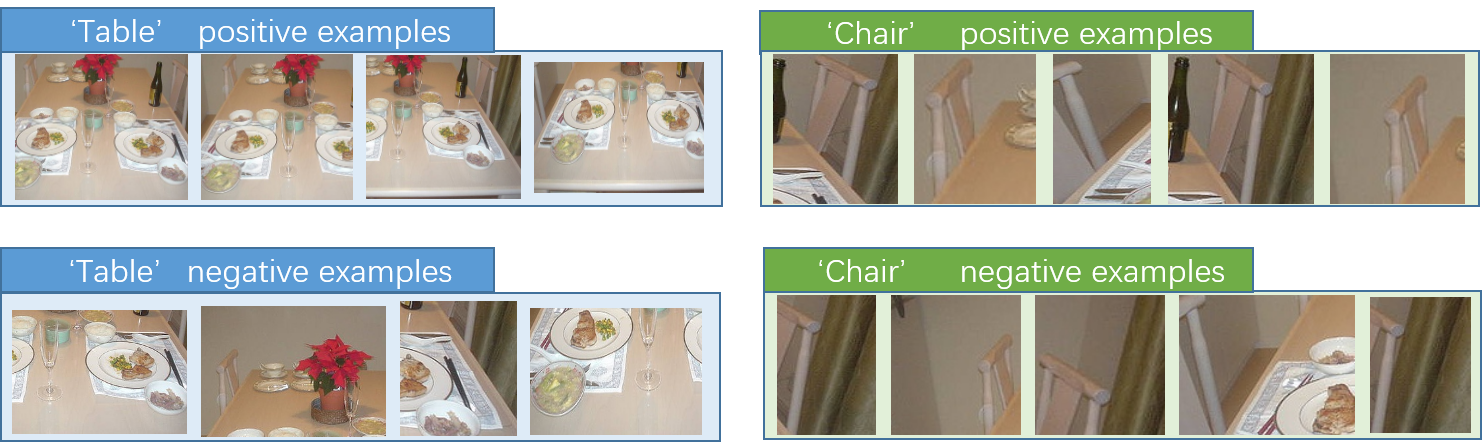}}
  \centerline{(c)}
\end{minipage}
\vspace*{-5pt}
\caption{Object detection is based on the classification of RoIs (proposals). The triplet embedding works on the groundtruth of object RoIs (a), the positive and negative RoIs (c) with respect to different object classes (like Table and Chair), in which the similarity distance constraint (b) is applied. }
\vspace*{-15pt}
\label{fig:example}
\end{figure*}

Moreover, some methods use bootstrapping (usually called hard examples mining) to improve performance. In \cite{girshick2014rich}, Ross \textit{el al.} proposed to cache all RoIs' features and employ a SVM classifier to identify hard examples and accordingly update detector models in an iterative manner. In \cite{online_hard_example_mining}, Shrivastava \textit{el al.} proposed to select hard negative RoIs online according to RoIs' classification and localization loss in the network forward stage, and then those RoIs' features are forwarded once again for better learning. In those methods, the rules of selecting hard examples are well-motivated towards effective learning. However, the distribution of positives vs. hard negatives in feature space, and their relative similarity distances are yet to be investigated to improve the classification performance in the context of object detectors.

In this paper, we propose triplet embedding to incorporate the constraint of relative similarity distances between positives vs. (hard) negatives into region-based detector learning. The learning object is to enforce that the similarity distance between any pair of RoIs from the same object class (positives) is smaller than the distance between any pair of RoIs from different classes including background negatives. Over FastRCNN~\cite{girshick2015fast} and OHEM~\cite{online_hard_example_mining} network models, we have elegantly implemented the triplet embedding.

Our contributions are twofold. First, to the best of our knowledge, this is the first work to incorporate triplet embedding into region-based detector learning, which has strengthened the classification of positives vs. (hard) negatives with respect to different object classes. Through jointly optimizing the injected triplet loss and the original loss with FastRCNN, we have significantly improved the detector performance. Second, we propose a so-called Top-K pooling to further improve detector performance, which is empirically shown to be effective in reducing noises in feature maps. The triplet embedding, together with Top-K pooling, has advanced the state-of-the-art FastRCNN and OHEM models. The superior performance has been demonstrated over benchmark PASCAL VOC2007 dataset.

The rest of this paper is organized as follows. In Section 2, we present the problem. In Section 3, we introduce the proposed approach. Comparison experiments are given in Section 4. Finally, we conclude this paper in Section 5.

\section{Problem Statement}

\label{sec:format}
Let's take the state-of-the-art FastRCNN as a baseline region-based detector, which is to optimize the joint objective of classification and localization, and performs end-to-end training in mini-batch stochastic gradient descent. Each true RoI from training images is labeled with a groundtruth class $g$ and a bounding-box regression target $t^*$. The optimization objective can be formulated as a multi-task loss:
\begin{equation}
\vspace{-5pt}
\begin{array}{cl}
L =L_{cls}(p,g) + 1 [ {g} \geq 1 ] L_{loc} (t,t^*),
\end{array}
\label{eq1}
\end{equation}
where $L_{cls}$ and $L_{loc}$ are the losses for classification and bounding-box regression, respectively. Specifically, $L_{cls}$ is a log loss and $L_{loc}$ is a smooth $L1$ loss. For training, $p$ is the predicted class label, and $t$ is the predicted box coordinates; $1 [ {g} \geq 1 ]$ equals $1$ when ${g} \geq 1$; otherwise it is background RoIs ($g=0$), then $L_{loc}$ is ignored.

From the classification point of view, those hard negatives lying close to decision hyperplanes are prone to misclassification. In the spirit of triplet embedding, it is beneficial to leverage similarity distance learning for better classification, in which those negatives are pushed away from positives, and meanwhile the positives of the same class are pulled together as shown in Fig.~\ref{fig:metric}. Hence, we aim to incorporate similarity distance constraint in multi-loss optimization:
\vspace{-2pt}
\begin{equation}
\begin{array}{cl}
D(R^g_i, R^n) >  D(R^g_i, R^p_j),
\end{array}
\vspace{-5pt}
\label{eq2}
\end{equation}
where $D$ is the similarity distance in feature space, $R^g_i$, $R^p_j$ and $R^n$ denote the groundtruth, positive and negative RoIs, respectively. Triplet embedding is then applied as shown in Fig.~\ref{fig:example} (b).

\section{Proposed Approach}

%\subsection{Embedding Triplet Loss  for Object Detection}
\begin{figure*}[!ht]
  \centering
  \vspace{-5pt}
  \includegraphics[width=1\linewidth]{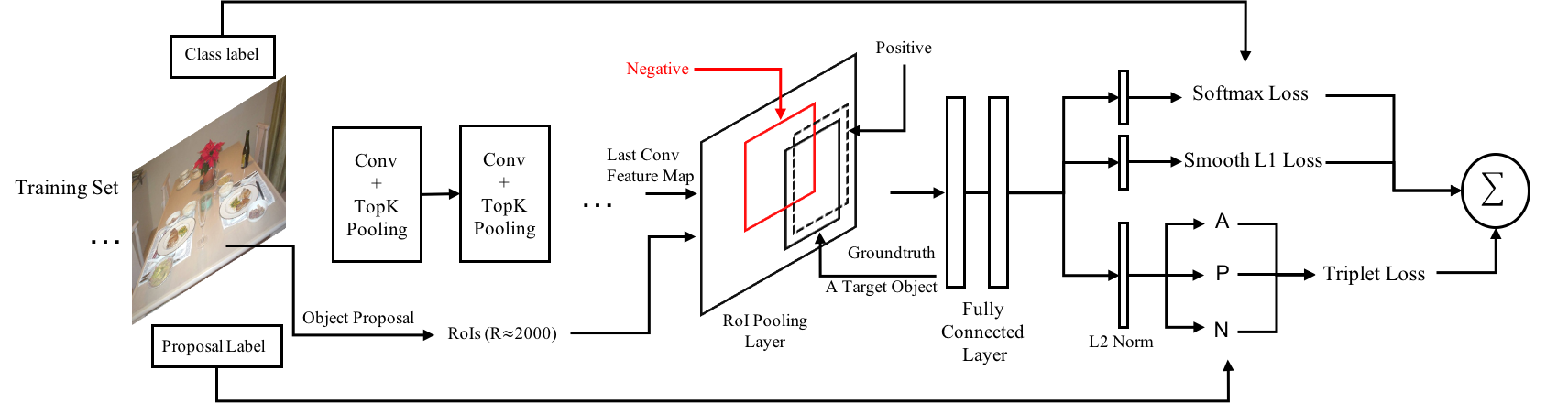}
  %\vspace*{-2pt}
  \caption{Illustration of a region-based object detection network with triplet embedding and Top-K pooling. In the forward stage, RoIs (object proposals) are generated and labelled with respect to different object/background classes. The triplet loss is added to strengthen the optimization objective and improve the classification performance consequently. }
 \label{fig:network}
  \vspace*{-10pt}
\end{figure*}
In this section, we first present RoIs samples selection for triplet embedding as well as hard triplet units, and then we describe joint optimization loss function.

\subsection{Incorporating Triplet Embedding}
The loss per RoI is the sum of a log loss and a smooth $L1$ loss in FastRCNN training. Here we consider a similarity distance constraint based on RoI's features. Specifically, given an image $X$ and a set of RoIs $(R_1,R_2,...,R_N)$ generated by object proposal algorithms as input to the network, we get the RoIs' features in the last fully connected layer $f(R)$. We use $L2~ norm$ distance to compute the similarity between RoIs as:
\begin{equation}
\begin{array}{cl}
D(R_i, R_j) = {\left\| f(R_i)-f(R_j) \right \|}^2_2.
\end{array}
\label{eq3}
\end{equation}
In network training, we setup the constraint of similarity distances as: the similarity distance $D(R_i^g,R_j^p)$ incurring object RoIs and their groundtruth are closer than the distance between the groundtruth and the background RoIs $D(R_i^g,R^n)$. Formally, this constraint can be formulated as:
\begin{equation}
\begin{array}{cl}
D(R^g_i, R^n) >  D(R^g_i, R^p_j)+\alpha,
\end{array}
\label{eq4}
\end{equation}
where $\alpha$ is the minimum margin between $D(R_i^g,R_j^p)$ and $D(R_i^g,R^n)$ as shown in Fig.~\ref{fig:metric}. We empirically set $\alpha=0.5$ in this work. Thus the loss of a triplet unit $<R_i^g, R_j^p,R^n>$ is defined as:
 \begin{equation}
\begin{array}{cl}
L(R_i^g, R_j^p,R^n) = \max(D(R_i^g,R_j^p) - D(R_i^g,R^n) + \alpha, 0).
\end{array}
\label{eq5}
\end{equation}
When the constraint in Eq (4) is violated for any RoIs triplet, the loss is back propagated. Therefore, the optimization objective is to minimize the loss function as:\
\begin{small}
 \begin{equation}
\begin{array}{cl}
L \!= \!\sum_{1}^{N}\max({\left \|f(R_i^g)\!-\!f(R_j^p)\right \|}^2_2 \!+\!\alpha\! -\! {\left \|f(R_i^g)\!-\!f(R^n)\right \|}^2_2, \!0),
\end{array}
\label{eq6}
\end{equation}
\end{small}
where $N$ is the total number of triplet units in training.

\subsection{Class-specific Triplet Embedding}
Each RoI (\textit{i.e.}, proposal) is assigned with a class label $l_{class}$ to indicate positive or negative RoI. The RoI labeling relates to the definition of foreground and background.

\textbf{Foreground RoIs.} A RoI is labeled as foreground when its IoU overlap with a groundtruth bounding box exceeds 0.5. Threshold 0.5 is compliant with the evaluation protocol in PASCAL VOC detection benchmark.

\textbf{Background RoIs.} A RoI is labeled as background when its maximum IoU overlap with any groundtruth bounding box falls into the interval $[bg_{low},0.5)$. $bg_{low}$ is the low overlap threshold, and in FastRCNN $bg_{low} = 0.1$ is setup, which supposes that those RoIs with very small overlap ($<0.1$) with any groundtruth bounding box are uncertain.

To effectively select negatives for triplet units, we introduce a class-specific label $l_{proposal}$ to each background (negative) RoI. Instead of a single category label, we propose to assign a class-specific label to each negative background RoI by using the class label of the groundtruth RoI with a maximum IoU overlap with the negative background RoI, as shown in Fig.~\ref{fig:example} (c). For example, $l_{proposal} = c$ means this negative RoI has a maximum overlap with an object of class $c$, and is likely to be a ``qualified" hard negative. As multiple object classes may be involved in an image, the RoIs are assigned to different groups $(G_1,G_2,...,G_M)$ according to $l_{class}$ and $l_{proposal}$, in which group $G_c$ consists of positive RoIs with $l_{class}=c$ and negative background RoIs with $l_{proposal}=c$. Accordingly, the triplet sampling strategy is applied between group-specific positives and background negatives. Referring to Eq (\ref{eq5})(\ref{eq6}), for each group $G_c$, $R_i^g$ is determined by the groundtruth, $R_j^p$ are RoIs with $l_{class}=c$ and $R^n$ are RoIs with $l_{class}=background$ and $l_{proposal}=c$.

\subsection{Hard Triplet Units Sampling}
The number of RoIs (proposals) produced by selective search \cite{selectivesearch} or edgebox \cite{edgebox} in an image is around $2000$, and the number of possible triplet units can reach up to $2000^3$. As a large portion of triplet units do not violate the similarity constraint, it is meaningful to select a subset of those hard triplet units to make effective and efficient training.

We select hard triplet units within $G_c$ through computing:
\begin{small}
\begin{equation}
\begin{array}{cl}
\text{argmax}_i{\left \| f(R_a^g)\!-\!f(R_i^p)\right \|}^2_2 \;\; and \;\; \text{argmin}_j{\left \| f(R_a^g)\!-\!f(R_j^n)\right \|}^2_2 \\
R_a^g, R_i^p, R_j^n \in G_c,
\end{array}
\label{eq7}
\end{equation}
\end{small}
where $R_a^g$, $R_i^p$ are groundtruth and positive RoIs and $R_j^n$ is negative RoI, $R_a^g$ works as a reference anchor that is selected from the groundtruth RoIs in the experiment.

\begin {table*}[!ht]
\label{tab:detection}
\scriptsize
\centering
\renewcommand\arraystretch{1.1}
\begin{tabular}{p{2.25cm}|p{0.85cm}|p{0.35cm}|p{0.22cm}p{0.22cm}p{0.22cm}p{0.23cm}p{0.22cm}p{0.21cm}p{0.21cm}p{0.21cm}p{0.23cm}p{0.23cm}p{0.23cm}p{0.23cm}p{0.23cm}p{0.23cm}p{0.23cm}p{0.23cm}p{0.23cm}p{0.23cm}p{0.22cm}p{0.21cm}}

\hline
Method &Train set&mAP& aero & bike & bird & boat & bottle & bus & car & cat & chair & cow & table & dog & horse & bike &persn &plant& sheep & sofa & train & tv\\
\hline
VGGM& 07& 59.4 & 71.6 & 71.9 & 55.9 & 51.8& 26.5 & 69.0 & 73.2 & 73.0 & 30.4 & 65.2 & 61.5 & 67.5 & 71.2 & 70.0 & 60.2 & 27.7 & 59.2 & 62.3 & 68.6 & 61.2\\
VGGM+TopK& 07& 60.0 & 71.5& 70.1 & 58.1 & 45.8 & 30.0 & 69.1 & 74.1 & 72.4 & 37.1 & 64.0 & 61.6 & 70.4 & 72.4 & 70.6 & 61.6 & 27.0 & 52.0 & 62.2 & 69.5 & 61.0 \\
VGGM+Triplet& 07& 61.1 & 70.3  & 73.0 & 57.9 & 48.2  & 29.6 & 67.4 & 73.2  & 72.9 & 39.3 & 66.4  & 63.0 & 68.1 & 71.6 & 71.3 & 63.0 & 28.8  & 58.2 & 62.7 & 73.0 & 64.1\\
VGGM+Triplet+TopK & 07& \textbf{61.6}& 70.5  & {73.5} & {58.7} & 49.3  & {31.6}  & 67.7 & {74.0} & 72.0 & {40.2} & {65.3}  & {62.1}  & {68.2}  & 71.2  & {72.3}  & {62.7} & {29.2}  & {59.5}  & {64.7} & {75.6} & 63.9\\
\hline
VGG16& 07& 66.9 & 74.5& 78.3& 69.2& 53.2& 36.6& 77.3& 78.2& 82.0& 40.7& 72.7& 67.9& 79.6& 79.2& 73.0& 69.0& 30.1& 65.4& 70.2& 75.8& 65.8 \\
VGG16+TopK&  07& 67.3 & 74.6 &  78.2 & 69.4 & 55.0 & 39.6 & 77.1  & 77.7 & 78.6  & 46.1  & 72.0 & 67.8  & 79.9  & 79.1  & 74.3 & 69.7  & 31.2  & 68.5  & 68.7  & 76.3  & 63.3\\
VGG16+Triplet& 07& 68.3 & 74.6  & 77.6  & 65.0  & 56.0  & 40.2  & 76.5  & 77.9  & 83.1  & 47.9  & 73.1  & 68.0  & 81.7  & 78.2  & 75.7  & 72.0  & 37.0  & 64.5  & 67.4  & 76.0  & 73.1\\
VGG16+Triplet+TopK & 07 & 68.7 & 75.6 & 78.6 & 68.0 & {56.2}  & 40.5 & {76.7}  & 78.9 & {84.3}  & 47.0 & 73.3 & {68.0}  &81.0  & {78.7}  & 75.4 & 72.2 & 37.8  & 65.5 & {67.8} & {76.5}  & {73.7}\\
MR-CNN \cite{multiregion} & 07 &\textbf{69.1} &\textbf{82.9} & 78.9 & 70.8 & 52.8 & \textbf{55.5} & 73.7 & 73.8 & 84.3 & 48.0 & 70.2 & 57.1 & 84.5 & 76.9 & \textbf{81.9} & 75.5 & 42.6 & 68.5 & 59.9 & 72.8 & 71.7 \\
Yuting et al.\cite{zhang2015improving}& 07 & 68.5 & 74.1 & \textbf{83.2} & 67.0 & 50.8 & 51.6 & 76.2 & 81.4 & 77.2 & 48.1 & 78.9 & 65.6 & 77.3 & 78.4 & 75.1 & 70.1 & 41.4 & 69.6 & 60.8 & 70.2 & 73.7 \\
\hline
VGG16 & 07+12 & 70.0 & 77.0  & 78.1  & 69.3  & 59.4  & 38.3  & 81.6  & 78.6  & 86.7  & 42.8  & 78.8  & 68.9  & 84.7  & 82.0  & 76.6  & 69.9  & 31.8  & 70.1  & 74.8  & 80.4  & 70.4 \\
VGG16+Triplet+TopK& 07+12& \textbf{72.1} & 79.0 & 78.8 & 71.9 & {62.0} & 42.7 & 80.0 & 80.5 &{87.2} & {48.5} & 80.3 &{72.1} & 83.4 & 84.8 & 77.2 & 71.3 &{39.9} & {72.5} & 73.9 & 83.2 & 72.6\\
AC-CNN\cite{attentivecontext} & 07+12 & 72.0 & 79.3 & 79.4 & 72.5 & 61.0 & 43.5 & 80.1 & 81.5 & 87.0 & 48.5 & 81.9 & 70.7 & 83.5 & \textbf{85.6} & 78.4 & 71.6 & 34.9 & 72.0 & 71.4 & \textbf{84.3} & 73.5\\
\hline
OHEM \cite{online_hard_example_mining} & 07 & 69.9& 71.2 & 78.3 & 69.2 & 57.9 & 46.5 & 81.8 & 79.1 & 83.2 & 47.9 & 76.2 & 68.9 & 83.2 & 80.8 & 75.8 & 72.7 & 39.9 & 67.5 & 66.2 & 75.6 & 75.9\\
OHEM \cite{online_hard_example_mining} + Ours & 07 & 71.7 & 74.4 & 80.9 & 72.1& 61.4 &49.7 & 80.9 &79.5 & 83.7 & 53.3 & 75.4& 71.4 & 80.7 & 81.9 & 76.8 & 74.8 & 42.4 & 68.5 & 73.1& 78.0 &75.1\\
OHEM \cite{online_hard_example_mining}&  07+12 & 74.6 & 77.7 & 81.2 & 74.1 & 64.2 & 50.2 & \textbf{86.2} & \textbf{83.8} & 88.1 & 55.2 & 80.9 & 73.8 & 85.1 & 82.6 & 77.8 & 74.9 & 43.7 & 76.1 & 74.2 & 82.3 & \textbf{79.6}\\
OHEM \cite{online_hard_example_mining} + Ours &07+12 & \textbf{75.8} & 79.6 & 81.7 & \textbf{75.2} & \textbf{66.4} & 54.7 & 84.0 & 83.1& \textbf{88.6} & \textbf{58.0} & \textbf{83.3} & \textbf{74.0} & \textbf{86.4} & 85.0 & 80.4 & \textbf{76.1} & \textbf{44.9} & \textbf{78.6} & \textbf{77.8} & 80.7 & 78.1\\

\hline
\end{tabular}
\caption{The detection performance comparisons over PASCAL VOC 2007. Different networks (VGGM, VGG16, OHEM) are applied. Separate results are given over two different training data: VOC 07 training set and VOC 07+12 training set, respectively. In addition, the impact of Triplet and Top-K pooling on detection performance is studied as well.}
\vspace*{-10pt}
\end{table*}

\subsection{Joint Optimization of Multiple Loss Functions}
Apart from the original classification and localization regression loss functions, we need to minimize the triplet loss, as illustrated in Fig.~\ref{fig:network}. Hence, a linear weighting is applied to form the optimization objective of multiple loss functions.
\begin{equation}
L_{total} = w_1 L_{cls} + w_2 L_{loc} + w_3 L_{triplet},
\end{equation}
where we empirically set $w_1 = 1$, $w_2 = 1$ and $w_3 = 0.5$ in this work. $L_{cls}$ and $L_{loc}$ are the classification and localization loss. $L_{triplet}$ enforces the similarity distance constraint. The L2 normalization is applied to the output of $fc7$ layer (last fully connected layer). The output of network contains : (1) a probability distribution over object classes and background, (2) regressed coordinates for bounding-box localization.

\section{Top-K Pooling}
\label{sec:pagestyle}

In the network forward stage, as layers go deeper, the size of feature maps become smaller and the noise influence in pooling operations would become more severe \cite{zhi2016two}. In this work, we propose a Top-K pooling to compute the mean of top K elements from sorting response values in the pooling window, which can be formulated as follows:
\begin{equation}
 \vspace{-5pt}
y_{i}= \frac {1}{K} \sum_{j=1}^{K} {x_{i,j}',}
\end{equation}
where $x_{i,j}$ denotes the $j_{th}$ element in $i_{th}$ pooling window and ${y_{i}}$ denotes the output of $i_{th}$ pooling window. ${x_{i,j}'}$ are elements of a sorted sequence in a descending order. For each $y_{i}$, a K-length array $R({y_{i}}) = \{x_{i, j} | j=1,2,...,K \}$ is maintained for the indices of the top K $x_{i,j}$ to readily compute the gradient.

Rather than applying Top-K pooling as post-processing in \cite{zhi2016two}, we not only compute Top-K responses in pooling windows, but elegantly incorporate pooling operation into network training. During the backward stage, the derivatives of the error at the layer's inputs are derived as:
\begin{equation}
\frac{ \partial E}{ \partial x_{i,j}} = \frac {1}{K} \frac{ \partial E}{ \partial y_{i}}, x_{i,j} \in R( y_{i}).
 \vspace{-3pt}
\end{equation}

Traditional max pooling is susceptible to noise, while Top-K pooling performs better than mean pooling in terms of capturing the statistics of response values. Note that top-K pooling may degenerate to max pooling or mean pooling when $K=1$ or $K=window\_size$.

\section{Experiments}
\label{sec:majhead}
\textbf{Datasets and Metrics}. We perform evaluation on Pascal VOC 2007 and 2012 datasets that contain 9963 and 22531 images, respectively. The datasets are divided into \textit{train}, \textit{val} and \textit{test} subsets, and contains 20 target object classes. The performance evaluation metric is mean of Average Precision (mAP). The overall performance and per-class performance are presented on VOC 2007 dataset.

\noindent\textbf{Implementation Details}. The deep learning platform Caffe \cite{jia2014caffe} is used to train the networks. Two FastRCNN models based on $VGG\_CNN\_M\_1024$ \textit{(VGG\_M)} and \textit{VGG16} network architectures are trained in the experiment. Both networks are initialized with the models pre-trained on ILSVRC 2012 image classification dataset. In addition, the input object proposals are provided by \textit{selective search} \cite{2013selective}.

\subsection{Experiment Results on VOC 2007}

Table 1 presents the detection results on VOC benchmarks. We firstly perform comparison experiments over $VGG\_M$. When incorporating both triplet loss and Top-K pooling into $VGG\_M$, we have achieved 2.2\% mAP improvements over 07 train\_val \textit{dataset}. Injecting triplet loss alone brings about 1.7\% mAP improvement, while Top-K pooling contributes to 0.6\% mAP improvement (from 59.4 to 60.0\% mAP). Over \textit{VGG16}, which is a deeper network than $VGG\_M$, we yield a mAP improvements of 1.8\% on 07 \textit{train\_val} dataset when combining the triplet loss and Top-K pooling. Under $07+12$ \textit{train\_val}, \textit{VGG16} has achieved up to 2.1\% mAP improvement. Moreover, compared to other typical region-based detectors, such as AC-CNN~\cite{attentivecontext}, Yuting~\cite{zhang2015improving}, MR-CNN~\cite{multiregion}, the proposed approach yields competitive performance as well. OHEM~\cite{online_hard_example_mining} is the state-of-the-art object detection approach, which has introduced online bootstrapping to the design of network structure based on the FastRCNN framework. As listed in Table 1, our method can further improve the detection performance of OHEM, and yields 1.2\% mAP improvements.

Note that the mAP improvements from Top-K pooling on \textit{VGG\_M} are more than on \textit{VGG16}. It may be attributed to the different sizes of pooling kernels. \textit{VGG\_M} adopts $3\times3$ kernel-size max pooling and length $2$ stride while \textit{VGG16's} default kernel-size is $2\times2$ and stride is $1$. For large max pooling kernel-size, the replacing with Top-K pooling can effectively mitigate the negative effects of background noise. However, although $2\times2$ pooling interval is very dense in \textit{VGG16}, $0.4$\% mAP gain still results with Top-K pooling ($K=2$).
\begin{figure}
  \centering
  \includegraphics[width=1\linewidth]{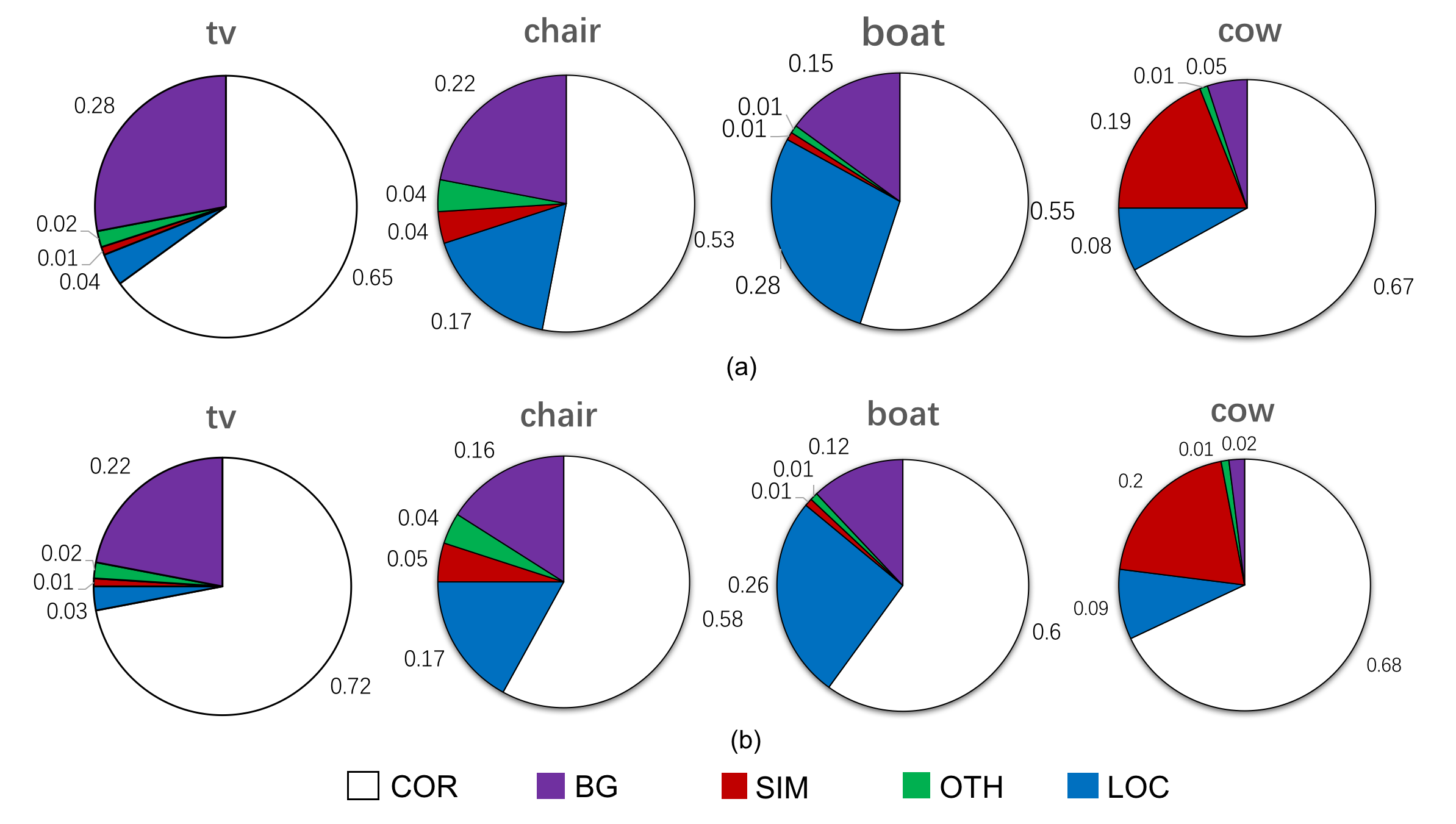}
  \vspace*{-5pt}
  \caption{The performance analysis of exemplar top detections with the original FastRCNN (a) and our approach of TripletLoss + Top-K pooling over FastRCNN (b). The pie charts show the percentages of Correct Detection (COR), False Positives from poor localization (LOC), visually similar objects (SIM), other VOC objects (OTH), or background (BG).}

  \label{fraction_analysis}
  \vspace{-10pt}
\end{figure}

As listed in Table 1, compared with FastRCNN, our method has achieved better detection results over most of object categories. In addition, we depict the pie charts with the percentages of true positives (correct) and false positives like \cite{hoiem2012diagnosing} in Fig.~\ref{fraction_analysis}. With triplet loss and Top-K pooling, the percentage of background false positives (BG) has been reduced. In particular, the performance improvements on some classes like tvmonitor, chair are significant. For instance, the mAP gains on chair reach up to 10\% in \textit{VGG\_M}. This can be partially attributed to similar visual characteristics of positive and negatives, while our enforced distance optimization is beneficial for more accurate discrimination. As to other classes like boat or cow, the detection performance is mainly affected by localization, so the improvements from our approach are limited.

\subsection{Discussion}
The state-of-the-art OHEM [13] applies the idea of online bootstrapping in designing network structure to improve learning. Both OHEM and our proposed approach attempt to deal with hard negative RoIs. It is worthy to note that our approach can be elegantly integrated into any region-based detector learning. For example, the triplet loss can be applied in OHEM as an additional optimization objective. From Table 1, the mAP improvements of OHEM with the proposed triplet embedding and Top-K pooling are 1.8\% and 1.2\% over 07 and 07+12 \textit{train\_val}, respectively.

To investigate the impact of triplet embedding and Top-K pooling on FastRCNN training, we illustrate the loss changes of SoftMax + Smooth L1 over VGGM network. As shown in Fig. 5, over the training course, the loss of original FastRCNN can be significantly reduced with our approach. The lower loss values have demonstrated the advantages to improve the efficiency and effectiveness of learning region-based detectors.

\begin{figure}
\centering
 \vspace{-5pt}
\includegraphics[width=0.95\linewidth]{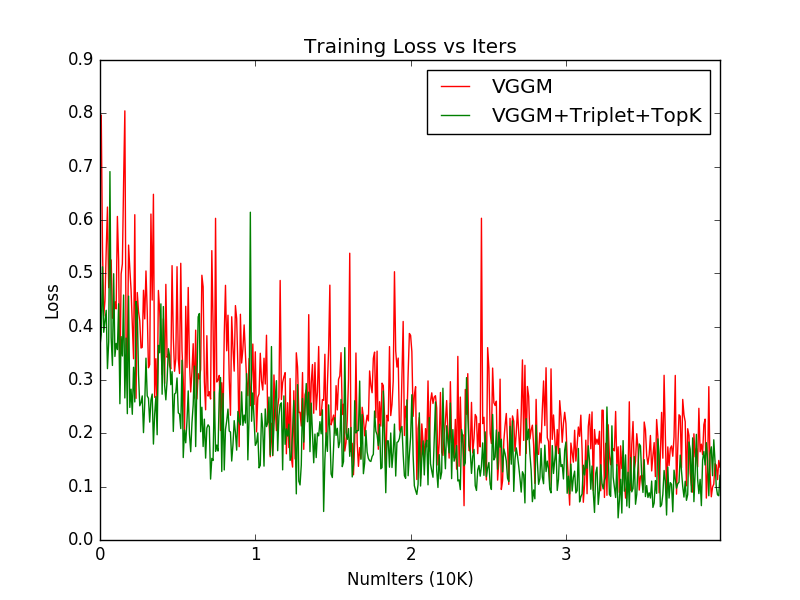}
 \vspace{-15pt}
\caption{Comparison of training loss (Softmax + Smooth L1) over VGGM with and without triplet loss + Top-K pooling. For fair comparison, although the additional triplet loss is applied in optimization objective, the shown loss values (of green curve) have deducted the portion of triplet loss.}
\label{fig:loss_curve}
 \vspace{-15pt}
\end{figure}
\section{Conclusion}

We have proposed to incorporate triplet embedding into the optimization objective of region-based detectors. The triplet loss may effectively enforce the similarity distance constraints between groundtruth, positive and negative RoIs. Moreover, a practically useful Top-K pooling is employed to further reduce the negative effects of feature map noises in network training. The proposed triplet embedding and Top-K pooling have significantly advanced the state-of-the-art FastRCNN and OHEM models, which have been demonstrated over PASCAL VOC 2007 benchmark.
%\section{Acknowledgments}
%\label{sec:concl}
\vspace{10pt}

\textbf{Acknowledgments:} This work was supported by grants from National Natural Science Foundation of China (U1611461, 61661146005, 61390515) and National Hightech R\&D Program of China (2015AA016302). This research is partially supported by the PKU-NTU Joint Research Institute, that is sponsored by a donation from the Ng Teng Fong Charitable Foundation.

\bibliographystyle{IEEEbib}
\bibliography{icme2017template}

\end{document}